\documentclass[11pt,a4paper]{article}
\usepackage[hyperref]{emnlp-ijcnlp-2019}
\usepackage{times}
\usepackage{latexsym}
\usepackage{natbib}
\usepackage{tikz}
\usepackage{pgfplots}
\usepackage{amsmath}
\usepackage{textcomp}
\usepackage[ruled]{algorithm2e}
\usepackage{multirow}
\usepackage{forest}
\usepackage{float}
\usepackage{comment}
\usepackage{xcolor}

\usetikzlibrary{shapes.geometric,arrows}

\DeclareMathOperator*{\argmax}{arg\,max}

\usepackage{url}

\newcommand{\dataset}[0]{\mathbf{D}}
\newcommand{\validset}[0]{\mathbf{V}}

\newcommand{\advising}[0]{Advsising\xspace}
\newcommand{\atis}[0]{ATIS\xspace}
\newcommand{\geo}[0]{GeoQuery\xspace}
\newcommand{\astbpe}[0]{AST BPE\xspace}

\newcommand{\fdkb}[0]{FD\&K baseline\xspace}
\newcommand{\dlseqtree}[0]{D\&L seq2tree\xspace}
\newcommand{\iyeretal}[0]{Iyer et al.\xspace}

\aclfinalcopy 

\title{Byte-Pair Encoding for Text-to-SQL Generation}

\author{Samuel M\"uller \\
  University of Cambridge  \\
  {\tt samuelgabrielmuller@gmail.com} \\\And
  Andreas Vlachos \\
  University of Cambridge  \\
  {\tt av308@cam.ac.uk} \\}

\date{}

\begin{document}
\maketitle
\begin{abstract}
Neural sequence-to-sequence models provide a competitive approach to the task of mapping a question in natural language to an SQL query, also referred to as text-to-SQL generation.
The Byte-Pair Encoding algorithm (BPE) has previously been used to improve machine translation (MT) between natural languages. 
%
In this work, we adapt BPE for text-to-SQL generation.
As the datasets for this task are rather small compared to MT,
we present a novel stopping criterion that prevents overfitting the BPE encoding to the training set. Additionally, we present \astbpe, which is a version of BPE that uses the Abstract Syntax Tree (AST) of the SQL statement to guide BPE merges and therefore produce BPE encodings that generalize better.
%
We improved the accuracy of a strong attentive seq2seq baseline on five out of six English text-to-SQL tasks while reducing training time by more than 50\% on four of them due to the shortened targets. Finally, on two of these tasks
we exceeded previously reported accuracies.
The implementation is available at \href{https://github.com/SamuelGabriel/sqlbpe}{https://github.com/SamuelGabriel/sqlbpe}.
\end{abstract}

\section{Introduction}
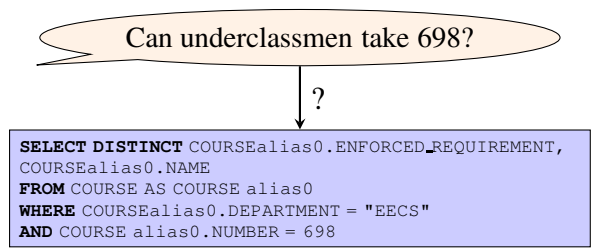
\begin{figure}
    \centering
    \tikzstyle{arrow} = [thick,->,>=stealth]
\begin{tikzpicture}[scale=0.4,font=\scriptsize]
\draw (0,0) node(b)[draw, text width=210, fill=blue!20]
{\texttt{\textbf{SELECT DISTINCT} COURSEalias0.ENFORCED\_REQUIREMENT, COURSEalias0.NAME\\
\textbf{FROM} COURSE AS COURSE alias0 \\
\textbf{WHERE} COURSEalias0.DEPARTMENT = "EECS"\\ 
\textbf{AND} COURSE alias0.NUMBER = 698}};
\draw (0,5) node(a)[draw, ellipse callout, callout relative pointer = {(6:-1)}, aspect = 3, fill=orange!10] {\normalsize Can underclassmen take 698?};
\draw [arrow] (a) -- (b) node[midway,right] {\large ?};
\end{tikzpicture}
    \caption{The task of text-to-SQL generation is to generate a query given a question in  natural language. This is an example from the Advising dataset \citep{finegan2018improving}.}
    \label{fig:example}
\end{figure}

Information is often stored in relational databases, but these databases can only be accessed with specialized programming languages, like the Structured Query Language (SQL). A natural language interface to a database (NLIDB) is a system that allows users to ask the database questions or give commands in natural language instead of using any programming language 
e.g.\ by mapping natural language to SQL queries like in the example in Figure \ref{fig:example}. Advances in text-to-SQL generation can also be relevant to other tasks concerned with the generation of other programming languages from natural language.

{}

Earlier approaches to solve this task were either rule-based \citep{popescu-etal-2004-modern} or based on statistical machine translation models \citep{Andreas2013SemanticPA}.
Recently, \citet{finegan2018improving} showed strong results using neural sequence-to-sequence models on English-language text-to-SQL task, in line with other recent work by \citep{Iyer_2017,Dong_2016},
even though these models have to predict rather long sequences when predicting SQL queries. 
For comparison, the translation dataset Multi30k \citep{elliott-EtAl:2016:VL16} for example has a mean target length that is less than a sixth of that of targets in the \advising \citep{finegan2018improving} dataset, one of the text-to-SQL datasets we evaluate on. The average length of the targets in the Multi30k dataset increases to over three quarters of the average length in \advising when predicting on the character-level. 
Interpreting the sentence as a sequence of characters gives the model more flexibility and allows it to predict previously unseen words, but the increased sequence length also takes its toll on predictive accuracy and speed \citep{Sennrich_2016}.
As a remedy, \citet{Sennrich_2016} proposed to apply the Byte-Pair Encoding algorithm (BPE) \citep{gagebpe} on target sequences for character-level machine translation. BPE is a compression algorithm that encodes commonly co-occurring characters into single symbols.

In this work we use BPE to compress token-level SQL targets to shorter sequences with a flexible stopping criterion and guidance by the abstract syntax tree (AST).
We improved the accuracy of a strong attentive seq2seq baseline on five out of six English text-to-SQL tasks while reducing training time by more than 50\% on four of them due to the shortened targets. Finally, on two of these tasks
we exceeded previously reported accuracies.
We are able to improve on strong baselines in accuracy, training time and inference time with our methods on most SQL-to-text tasks.


\section{Background}
In the text-to-SQL task we want to learn a model $p$ of SQL queries of the form $a=y_1\dots y_{|a|}$ conditioned on natural language input of the form $q=x_1\dots x_{|q|}$. 
We want to find a $p$ such that our accuracy is high on the test set. We approximate this goal by search for a $p$ from some class of models that is as close as possible to $\argmax_p \prod_{(q,a) \in T}p(a|q)$,
where $T$ is the test set. 
For this purpose we use the sequence-to-sequence neural network approach \citep{sutskever2014sequence} which constructs the SQL query incrementally, i.e.\ $p(a|q) = \prod_{t=1}^{|a|}p(y_t|y_{1:t-1},q)$. In this framework the question $q$ is encoded and decoded with a Long Short-Term Memory unit (LSTM) \citet{hochreiter1997long}. Additionally we employ an attention mechanism \citep{Bahdanau2014NeuralMT} to allow the decoder to flexibly combine the states of the encoder.

\section{BPE for text to SQL generation}
\label{methods}
While \citet{Sennrich_2016} applied BPE \citep{gagebpe} on characters inside words, we apply BPE on tokens, since the vocabulary of SQL queries is quite restricted 
(ignoring named entities); e.g.\ the anonymized \advising dataset  contains only 177 unique tokens. 
BPE is a compression algorithm that groups commonly co-occurring symbols into single symbols. BPE works by iterating over a dataset in $k$ scans, where $k$ is given. In each scan, the most frequent pair of consecutive symbols is replaced with a new symbol. This way, frequent co-occurring sequences of symbols can be predicted in one step, simplifying the task.
In this paper, we apply BPE to tokenized queries by interpreting each token as a symbol and combining neighboring tokens as pairs to create new tokens. 

\begin{figure}[]
\centering
	\begin{tikzpicture}[scale=0.4,font=\tiny]
	\foreach \t/\i in {SELECT/0,NAME/3,FROM/6,CITY/9,WHERE/12,STATE/15}
		\draw (\i, 0) node[below](b\i) {\t};
	\draw (17, 0) node[below,yshift=-0.58ex](b17) {. . .};
	\draw (b0) -- (1.5,2) node(a)[above] {SELCT NAME} -- (b3);
	\draw (b12) -- (13.5,2) coordinate(c) -- (b15);
	\draw (b9) -- (12,4) node[above](d) {CITY WHERE STATE} -- (c);
\end{tikzpicture}
    
\caption{This figure shows the encoding of an example query in part (a) and the applied BPE rules in part (b).}
	\label{fig:vanillabpe}
\end{figure}
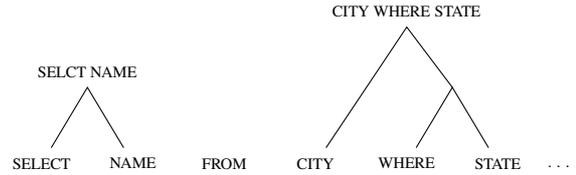

Figure \ref{fig:vanillabpe} illustrates an application of BPE encoding, where each node represents one token. Throughout the paper we use the training and validation sets to build and tune the BPE encoding on to ensure that the models developed are fairly assessed on the test sets.
To encode a new dataset with a given list $l$ of BPE encoded entries, one follows the same procedure as for the generation of the BPE encoding, but just applying rules found previously instead of creating new ones.

\newcommand\mycommfont[1]{\footnotesize\color{black!70}{#1}}
\SetCommentSty{mycommfont}
\begin{algorithm}[h]
\footnotesize
\SetAlgoLined
\caption{BPE with stopping criterion}
\KwIn{A training set $\dataset$ of target queries, and a validation set $\validset$ of target queries. The number of retention steps $r$ and the minimum number of occurrences in the training set for tokens appearing in the validation set $m$.}
\KwOut{A list l of pairs of tokens representing a BPE encoding as well as BPE encoded versions of $\dataset$ and $\validset$.}
\SetArgSty{textnormal}
l = \texttt{EmptyList}()\;
\While{\text{diff $<$ r and $\dataset$ contains a bigram}}{
maxpair = \texttt{pairWithMaxCount}($\dataset$)\tcp*{maxpair is the most common token pair in $\dataset$.}
newoov = \texttt{addsNewOOV}($\dataset$, $\validset$, maxpair, m)\;
\uIf{newoov}{
    diff += 1\;
}\uElse{
    l.append(maxpair)\;
    
    $\dataset$, $\validset$ = \texttt{replacePair}($\dataset$, $\validset$, maxpair)\tcp*{Replace occurences of maxpair with a new token.}
}}
\SetKwFunction{proc}{addsNewOOV}
\SetKwProg{myproc}{Procedure}{}{}
\SetArgSty{textnormal}
\myproc{\proc{$\dataset$, $\validset$, $\text{pair}$, c}}{
oovcount = $|\texttt{vocabulary}(\validset, 1) \backslash \texttt{vocabulary}(\dataset, \text{c}) |$\tcp*{vocabulary returns the set of tokens with a minimum number of occurences in a dataset.}
$\bar\dataset$, $\bar\validset$ = \texttt{replacePair}($\dataset$, $\validset$, pair)\;
newoovcount = $|\texttt{vocabulary}(\bar\validset, 1) \backslash \texttt{vocabulary}(\bar\dataset, \text{c}) |$\;
\Return newoovcount $>$ oovcount\;
}
\label{bpestoppingcriterion}
\end{algorithm}

\section{A stopping criterion for BPE}
\label{section:bpestoppingcriterion}
In previous work the number of BPE scans $k$ was treated as a hyper-parameter that needs to be hand tuned \citep{Sennrich_2016}. While this might work for tasks where the performance of the model is robust against different values of $k$, we found that this is not the case for text-to-SQL generation, due to the small size of the datasets. 
%
%
If $k$ is set too low, we do not experience all the benefits of BPE, since we could shorten sequences further. If $k$ on the other hand is set too high, there is a risk that our model is unable to predict some combinations of tokens.
This situation might arise, for example, if in the training set a token $a$ is always followed by a token $b$, and therefore the model combines these two tokens into a new token $x$. Since all occurrences of $a$ are followed by $b$, applying a BPE step will remove $a$ completely from the dataset. Therefore if there is a sequence in the test data that requires generating $a$ without $b$ following it the model would not be able to.

To ameliorate this issue we propose a stopping criterion for BPE as outlined in Algorithm \ref{bpestoppingcriterion}, which has two less sensitive hyper-parameters, $r$ and $m$, instead of the number of steps $k$. We were able to use the same settings for these new parameters on many different datasets and tasks with competitive results, which preliminary experiments showed not to be possible with a fixed number of steps $k$.
The method is outlined in algorithm \ref{bpestoppingcriterion}. We keep track of all tokens present in the training and the validation set as we apply consecutive BPE steps and stop as soon as we took $r$ steps that leave tokens in the validation set that can\textquotesingle{t} be found in the training set no more. 
The second parameter $m$ is the minimum number of occurrences in the training set for each token in the validation set. This is equivalent to ensuring that a minimum count is fulfilled for each token added to the vocabulary with BPE.

\section{AST BPE}
It was previously shown that it can be helpful to consider the abstract syntax tree (AST) of SQL queries for query generation \citep{dong2018coarse}. Similar to the AST, the BPE algorithm defines a tree structure onto queries, but it might not be well aligned with the query's AST. 

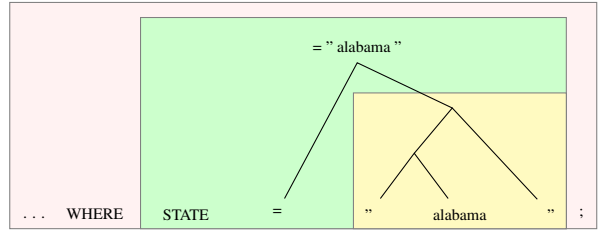
\begin{figure}[t!]
    \center
    \def\la{1.5}
\def\lb{3}
\def\lc{4.5}
\def\bb{3.5}
\def\bc{\lc+1.5}
\def\bd{\bc+.5}
\def\semi{28}
\begin{tikzpicture}[scale=0.4,font=\tiny]
\filldraw[gray, fill=pink!20!white] (9.2,-1) rectangle (\semi+0.5,\bd);
\filldraw[gray, fill=green!20!white] (13.5,-1) rectangle (27.5,\bc);
\filldraw[gray, fill=yellow!30!white] (20.5,-1) rectangle (27.5,\bb);
	\draw (10, 0) node[below,yshift=-0.58ex](b10) {. . .};
	\foreach \t/\i in {{WHERE}/12,STATE/15,=/18,"/21,alabama/24,"/27,;/\semi}
		\draw (\i, 0) node[below](b\i) {\t};
	\draw (b21) -- (22.5,\la) coordinate(e) -- (b24);
	\draw (e) -- (23.75,\lb) coordinate[above](g) -- (b27);
	\draw (b18) -- (20.625,\lc) node(h)[above] {= " alabama "} -- (g);
\end{tikzpicture}
\caption{Illustration of an example query\textquotesingle{s} AST BPE representation. Boxes of different color illustrate different levels in the AST.}
\label{fig:astbpe}
\end{figure}

The idea of AST BPE is to keep the main principle of BPE, but align the BPE structure 
with the AST by restricting what is interpreted as a pair when computing a BPE encoding to sub-sequences that are aligned in the AST.
More formally, consider two tokens $a$ and $b$, that represent the token sequences $a_1,\dots,a_{|a|}$ and $b_1,\dots,b_{|b|}$ respectively, in the dataset $\dataset$ after a number of BPE steps. In the \astbpe setup the PairCounter in Algorithm \ref{bpestoppingcriterion} only considers $a$ and $b$ as neighbors if the sequence built by concatenating the two represents a set of neighboring sibling AST nodes.
An example query encoded with \astbpe can be found in Figure \ref{fig:astbpe}. The colored boxes illustrate the levels of the query\textquotesingle{s} AST.
This method is especially helpful for the small datasets found in the text-to-SQL domain, since on larger ones that represent the target distribution well, vanilla BPE is likely to chooses tokens in a way such that are aligned with the AST. 
On small datasets like \geo on the other hand, we could see that vanilla BPE for example encodes closing parentheses and following keywords as BPE tokens.

\section{Related Work}
\label{relatedwork}
As far as we know there is no previous work on the application of BPE to the text-to-SQL task or any other structured language generation task. There exists research on related topics though.
\citet{dong2018coarse} explored how to use a flexible form of templates by using a neural sequence-to-sequence model which generates targets that only contain a coarse representation of the query that is filled with entities and identifiers by a second model. In contrast to BPE this model increases the complexity of the approach as two models need to be trained. 
\citet{finegan2018improving} proposed a baseline model that is only based on templates, which are not dynamically predicted but gathered from the  training data. Based on the representations of queries in the anonymized version of a dataset, they built an index of distinct queries (up to entity names). Then, they used an LSTM-based classifier to classify a question as one of the anonymized queries in the training set and classify input tokens as entities that replace placeholders. This technique only yields a simple baseline, since it does not generalize to unseen queries. We refer to this method as the \fdkb. 
The AST of SQL queries was previously used to predict queries by \citet{finegan2018improving}, who applied the methods \citet{Dong_2016} developed for logical parsing to SQL. They generate the query recursively over the AST, while we use a sequential representation of the query at prediction time and just use tree structures over the queries to find common parts of queries to unify. We refer to the adaptation of this method to SQL by \citet{finegan2018improving} as \dlseqtree.
\citet{Iyer_2017}, similar to \citet{Jia_2016} used the tree structure inherent to logical forms and artificial languages to grow their datasets and therefore also implicitly teach the model the modularity and replaceability of nodes in the parse tree. In the following we will call this method \iyeretal 

\section{Results}

We report our results for both BPE and \astbpe with an attention-enabled seq2seq model for all datasets and report the accuracy of predicting the whole query exactly as the target query.
We evaluate on the \advising dataset, which counts 4570 questions, as well as the simultaneously re-published datasets \atis \citep{finegan2018improving,data-atis-original,data-atis-geography-scholar}, an air traffic related dataset of 5280 questions, and \geo \citep{finegan2018improving,data-geography-original,data-atis-geography-scholar}, a dataset regarding the geography of the United States of America of 877 questions. All datasets have English as their natural language part.

\newcommand{\twoline}[2]{\begin{tabular}{@{}c@{}}#1 \\ #2\end{tabular}}

\newcommand{\twolineleft}[2]{\begin{tabular}{@{}l@{}}#1 \\ #2\end{tabular}}

\newcommand{\timecell}[2]{\tiny \twolineleft{#1}{#2}}

Our evaluation encompasses experiments on two split; one where the datasets in train, validation and test set are split based on the questions asked, which have examples with the same target across these sets, and one split based on the query, where each query is only contained in one of the sets.
We used the same hyper-parameters for both dataset splits. For all experiments the retention steps parameter $r$ was set to 20 and the minimum frequency in the training set $m$ was set to 100 for all datasets, besides \advising for which it was set to 300.
We ran all experiments on a single Nvidia Tesla M60 GPU. 
In table \ref{tab:results} our results on the test sets can be found.

\paragraph{Training Details}
All models use bidirectional LSTMs for encoding, with a hidden state size of 100. We initialize the LSTMs for encoding with zeroed hidden states. For decoding we use a LSTM and treat the initial hidden state as a parameter. We used concurrently-trained token embeddings of size 100 for all models.
To extract the AST from the queries in the training set for \astbpe we use the Python library `sqlparse' by Andi Albrecht.\footnote{\url{https://github.com/andialbrecht/sqlparse}}

During training we applied a dropout of 0.5 on the input and output of the LSTMs. We used batches of size 32 and the Adam optimizer \citep{kingma2014adam} with a constant learning rate of 1e-3. All weights were initialized with the default PyTorch weight initialization \citep{paszke2017automatic}. At inference time beam search with a beam width of 3 was applied. We employed early stopping guided by the validation set and a retention period of 50 epochs.

\begin{table*}[h]
\footnotesize
    \centering
    \begin{tabular}{c|c|l@{}ll@{}ll@{}l|l@{}ll@{}ll@{}l}
        && \multicolumn{6}{c|}{Question Split}& \multicolumn{6}{c}{Query Split}\\
    	Model & BPE Usage & \multicolumn{2}{c}{\advising} & \multicolumn{2}{c}{\atis} & \multicolumn{2}{c|}{\geo} & \multicolumn{2}{c}{\advising} & \multicolumn{2}{c}{\atis} & \multicolumn{2}{c}{\geo}\\\hline\hline
    	Seq2seq & \multirow{2}{*}{No BPE} &79.93\% & \timecell{9h12m}{3m43s} & 51.68\% & \timecell{28h20m}{5m47s} & 59.14\% & \timecell{1h11m}{1m47s}  & 0.00\% & \timecell{3h43m}{8m38s} & 8.65\% & \timecell{6h31m}{4m44s} & 0.00\% & \timecell{1h58m}{0m25s}\\\cline{1-1}
    	\multirow{3}{*}{\twoline{Seq2seq}{with attention}} && \textbf{89.01}\% & \timecell{6h8m}{4m36s} & 56.60\% & \timecell{22h40m}{6m29s} & 70.25\% & \timecell{1h36m}{1m1s}&3.66\% & \timecell{4h27m}{11m38s} & 25.94\% & \timecell{11h16m}{5m51s} & 47.25\% & \timecell{2h43m}{1m36s}\\\cline{2-2}
    	& BPE & 88.13\% & \timecell{3h56m}{2m18s} & 57.05\% & \timecell{11h51m}{4m16s} & 70.61\% & \timecell{24m18s}{1m45s}& 3.06\% & \timecell{3h45m}{6m24s} & 6.05\% & \timecell{2h41m}{2m36s} & 39.01\% & \timecell{1h21m}{0m24s}\\\cline{2-2}
    	& AST BPE & 87.61\% & \timecell{3h19m}{3m34s} & 56.38\% & \timecell{12h31m}{4m25s} & \textbf{72.40}\% & \timecell{31m5s}{1m46s}& 3.71\% & \timecell{5h1m}{6m28s} & 24.50\% & \timecell{4h35m}{3m43s} & \textbf{50.00}\% & \timecell{1h24m}{0m28s}\\\hline
    	\fdkb & \multirow{3}{*}{No BPE} & \multicolumn{2}{l}{89\%} & \multicolumn{2}{l}{56\%} & \multicolumn{2}{l|}{56\%}& \multicolumn{2}{l}{0\%} & \multicolumn{2}{l}{0\%} & \multicolumn{2}{l}{0\%}\\\cline{1-1}
    	\dlseqtree &  & \multicolumn{2}{l}{88\%} & \multicolumn{2}{l}{56\%} & \multicolumn{2}{l|}{68\%}& \multicolumn{2}{l}{\textbf{8\%}} & \multicolumn{2}{l}{\textbf{34\%}} & \multicolumn{2}{l}{23\%}\\\cline{1-1}
    	\iyeretal && \multicolumn{2}{l}{88\%} & \multicolumn{2}{l}{\textbf{58\%}} & \multicolumn{2}{l|}{71\%}& \multicolumn{2}{l}{6\%} & \multicolumn{2}{l}{32\%} & \multicolumn{2}{l}{49\%}\\
    \end{tabular}
    \caption{Accuracy alongside training (the upper time) and testing time (the lower time) on the datasets. Accuracy is measured on the test set. The results below the line are from \citet{finegan2018improving}.}
    \label{tab:results}
\end{table*}{}
\paragraph{Evaluation on the question split}
On the question splits BPE outperforms the attentive sequence-to-sequence for both \atis and \geo in both accuracy and training time, while especially \iyeretal could gain further improvements. These could be combined with both versions of BPE though.
On \geo, the \astbpe model outperforms the attentive sequence-to-sequence model by over 2\% in absolute accuracy, establishing a new state-of-the-art, and requiring only about a third of the time to train.

\advising 
is the only dataset on which we could not improve performance with BPE. 
Since \advising\textquotesingle{s} question split is structured such that each query in the question split of \advising appears in the training, the test and the validation set, a BPE encoding with a minimum count $m$ of 1 and 1 retention step $r$ can reduce the whole task to a classification task. BPE with these setting achieves an accuracy of 90.40\% with a training time of 48 minutes and a inference time of 15 seconds. It is worth noting that this setup surpasses even the \fdkb, although the only difference to it is that our model has an attention mechanism, for a simple classification. 

To further investigate the reasons for the good results with BPE methods across the datasets we took a closer look at the performance of \astbpe on \geo for unseen and seen queries. A query is \textit{seen}, if it is contained in the training data with a different question. Table \ref{analysegeo} shows that for the \geo task \astbpe did actually not improve accuracy on unseen queries, but instead improved the accuracy on seen queries, which make up 77\% of the test set.

\begin{table}[h]
	\small
	\centering
	\begin{tabular}{c|ccc}
	Query type & Seen queries & Unseen queries  \\\hline\hline
	No BPE  & 84.26\%  & 22.22\% \\
	\astbpe & 88.43\% & 17.46\% 	
	\end{tabular}
	\caption{Accuracy on \geo for queries (not) in the training set.}
	\label{analysegeo}
\end{table}
\paragraph{Evaluation on the query split}
We could improve performance on the query splits of both \advising and \geo with \astbpe. For \geo we set a new state-of-the-art on this task, and at the same time halve inference and test time compared to the base model. For \advising, the training time did not improve, even though \astbpe reduces the average query length in the training set by over 44\%; the effect of this could be seen in improved inference speed.

Only for \atis the BPE models did not improve accuracy, but training time could be more than halved at an absolute accuracy loss of less than 1.5\% with \astbpe.
A likely reason for why BPE did not improve accuracy on \atis, is that \atis contains many different query patterns, a pattern being a query type abstracted away from the table schema. Each pattern in \atis only appears in 7 queries on average. In the test set of \atis{\textquotesingle{}} query split over 47.84\% of the queries therefore have an unseen pattern, while on the query split of \geo for example only 5.49\% of queries have an unseen pattern. 

\begin{table}[h]
	\small
	\centering
	\begin{tabular}{c|cc}
	BPE setting & Unseen Patterns & Seen Patterns\\\hline\hline
	No BPE & 25.90\% & 23.20\%\\
	BPE & 1.20\% & 11.05\%\\
	\astbpe & 24.70\% & 25.41\%\\
	\end{tabular}
	\caption{This table shows the accuracy of seq2seq models with attention and different kinds of BPE on the queries from the test set with patterns (not) contained in the dataset. It can be seen that BPE and \astbpe are especially strong in predicting queries with a known pattern.}
	\label{table:query_atis_seen_unseen}
\end{table}

In Table \ref{table:query_atis_seen_unseen} we analyze the accuracy of different models on the queries with seen and unseen patterns on \atis. We can see that for BPE, which performs overall worst on this dataset, the performance is even worse for unseen patterns. For \astbpe the outcome is somewhat similar,
as it improves performance on seen patterns, but the performance on unseen patterns degrades.
This result aligns well with what we saw in Table \ref{analysegeo} for \astbpe on the question split of \geo, where we could see that \astbpe helped with seen queries, but not with unseen ones. 

\section{Conclusion}
\label{conclusion}
In this work we showed that BPE can be applied to text-to-SQL generation.
In particular we found that for anonymized datasets BPE was able to improve upon the base models in five out of six cases, and additionally cut the training time by more than 50\% in four of the cases. We showed that \astbpe is especially helpful for datasets split by query, which require the model to generalize to previously unseen queries and query structures. We could also observe that the biggest impact was on experiments with the smallest dataset, \geo, where we achieved new state-of-the-art results for both dataset splits.
The BPE methods developed in this work are not specific to SQL and could be applied to many other tasks requiring structured language generation.

\section*{Acknowledgments}
Andreas Vlachos is supported by the EPSRC grant eNeMILP (EP/R021643/1).

\bibliography{main.bib}
\bibliographystyle{acl_natbib}

\end{document}